\documentclass[graybox]{svmult}

\usepackage{type1cm}        
\usepackage{makeidx}         
\usepackage{graphicx}        
\usepackage{multicol}        
\usepackage{multirow}        
\usepackage[bottom]{footmisc}

\usepackage{newtxtext}       %
\usepackage{newtxmath}       

\usepackage{color}
\usepackage{subfig}

\usepackage{framed}

\usepackage{algorithm} 
\usepackage{algorithmic}

\usepackage[bottom]{footmisc}

\usepackage{array}
\newcolumntype{L}[1]{>{\raggedright\let\newline\\\arraybackslash\hspace{0pt}}m{#1}}
\newcolumntype{C}[1]{>{\centering\let\newline\\\arraybackslash\hspace{0pt}}m{#1}}
\newcolumntype{R}[1]{>{\raggedleft\let\newline\\\arraybackslash\hspace{0pt}}m{#1}}

\usepackage{xcolor}
\usepackage{graphicx}
\usepackage{colortbl}
\usepackage{array}
\usepackage{pifont}

\usepackage{tikz-qtree}

\definecolor{tabred}{RGB}{230,36,0}%
\definecolor{tabgreen}{RGB}{0,116,21}%
\definecolor{taborange}{RGB}{255,124,0}%
\definecolor{tabbrown}{RGB}{171,70,0}%
\definecolor{tabyellow}{RGB}{255,253,169}%
\newcommand*{\greenbullet}{\textcolor{tabgreen}{\ding{108}}}%

\makeindex             

\begin{document}

\title*{A brief overview of swarm intelligence-based algorithms for numerical association rule mining}
\titlerunning{A brief overview of SI-based algorithms for NARM}        
\author{Iztok Fister Jr. and Iztok Fister}

\institute{Iztok Fister Jr. \at University of Maribor, Faculty of Electrical Engineering and Computer Science, Koro\v{s}ka cesta 46, Slovenia, \email{iztok.fister1@um.si}
\and Iztok Fister \at University of Maribor, Faculty of Electrical Engineering and Computer Science, Koro\v{s}ka cesta 46, Slovenia, \email{iztok.fister@um.si}}

\maketitle

\abstract*{A}
Numerical Association Rule Mining is a popular variant of Association Rule Mining, where numerical attributes are handled without discretization. This means that the algorithms for dealing with this problem can operate directly, not only with categorical, but also with numerical attributes. Until recently, a big portion of these algorithms were based on a stochastic nature-inspired population-based paradigm. As a result, evolutionary and swarm intelligence-based algorithms showed  big efficiency for dealing with the problem. In line with this, the main mission of this chapter is to make a historical overview of swarm intelligence-based algorithms for Numerical Association Rule Mining, as well as to present the main features of these algorithms for the observed problem. A taxonomy of the algorithms was proposed on the basis of the applied features found in this overview. Challenges, waiting in the future, finish this paper.

\section{Introduction}
We live in a highly competitive world, where a lot of decision-making formerly carried out by humans are supported by  Artificial Intelligence (AI)~\cite{russell2002artificial}. AI has been changing the world in all domains, from advertisements to driverless cars~\cite{lipson2016driverless}. AI cannot exist without data that actually play a  similar role as an oil by driving a motor. However, decisions produced by the AI systems are based mostly on past data from the domain of interest. Data entering in the decision process can be in various forms, i. e. from unstructured, through semi-structured to structured data~\cite{rusu2013converting}. The AI methods applying to all these forms of data can discover new insights (i.e., knowledge) hidden in the data. Recent trends suppose that the AI solutions have still been rising and striving in each corner of human life~\cite{tegmark2017life}. Indeed, there are also some speculations about the superintelligence of the machines~\cite{chalmers2009singularity}.

\begin{figure*}
    \includegraphics[width=1\textwidth]{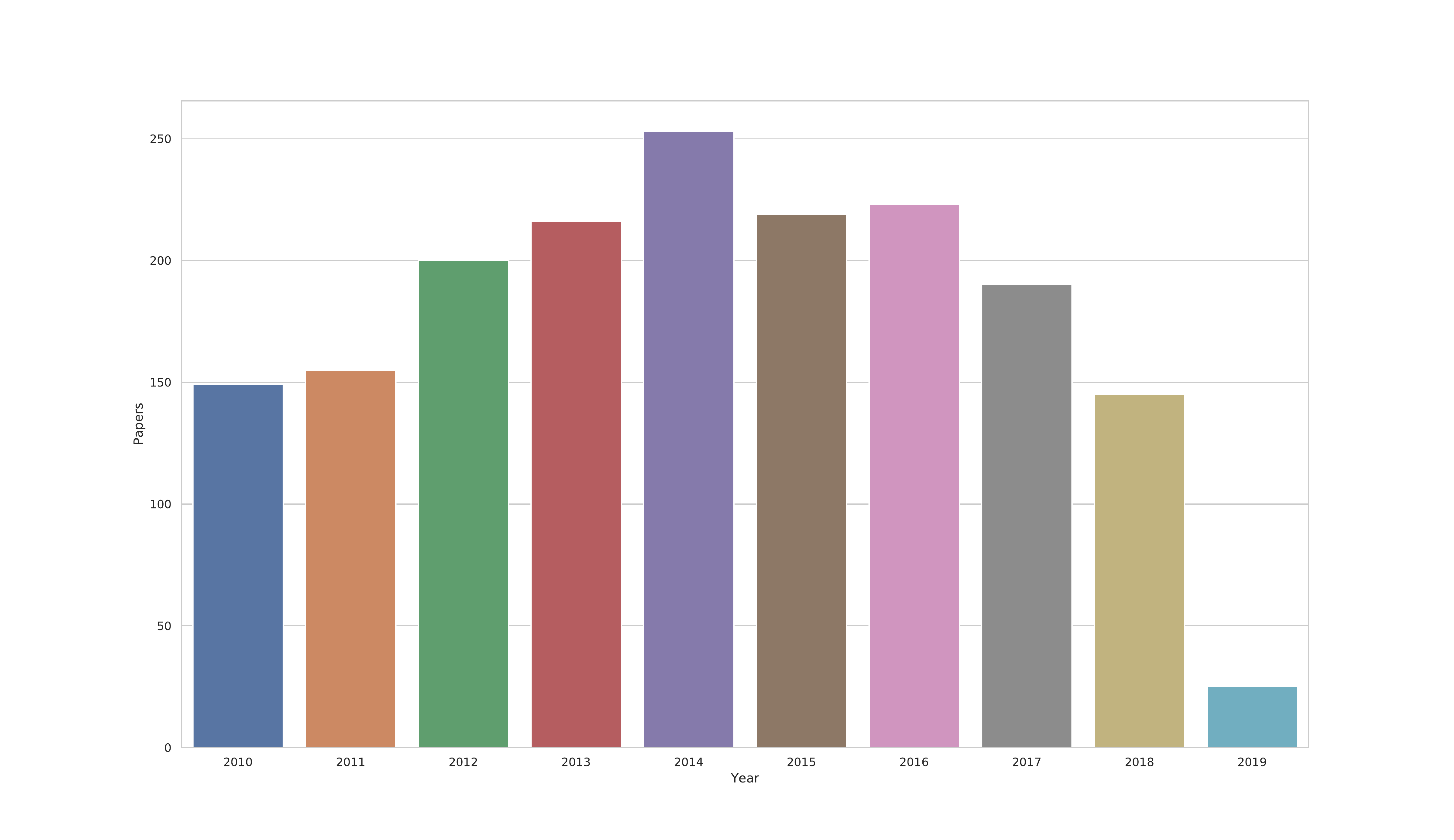}
    \caption{The number of papers including ARM during the last decade.}
    \label{evolution-arm}
\end{figure*}

A research area which utilizes the AI methods and data is called Machine Learning (ML)~\cite{alpaydin2016machine}, where researchers are occupying themselves with different methods under the ML umbrella. Nowadays, a lot of data are stored in transaction databases structured as a matrix with rows and columns, where each instance comprises of one row and each column denotes an attribute. Such structured data offer a huge opportunities for data mining methods, where Association Rule Mining (ARM) holds an important place. The task of ARM is to find relationships between attributes in a transaction database. These relationships are presented as  implications, where the left side of the mined rule represents an antecedent and the  right side the consequent. Decades of research and development in this field have revealed that the ARM is  very appropriate for application to market basket analysis~\cite{agarwal1994fast}, to building of intelligent systems~\cite{fister2019computational}, to development of rule-based classifiers~\cite{liu1998integrating}. Fig.~\ref{evolution-arm} depicts the frequency of papers which contains the term "association rule mining" in the title during the last decade. The frequencies were extracted using Microsoft Academic Graph~\cite{sinha2015overview}. 

As can be seen from the Figure, a zenith was achieved by this method in 2014. After this year, the interest decreased regarding it. Despite the maturity of this research area, this method is still very popular among researchers. Every year many new methods from the ARM field have  appeared, along with different real-world applications~\cite{altay2019performance}.

Numerical Association Rule Mining (NARM)~\cite{altay2019performance,Fister2018ARM} basically extends the idea of original ARM operating with categorical attributes only, where the algorithms for NARM are also able to mine the association rules within datasets also consisting of numeric attributes without discretization. Typically, algorithms designed for solving these problems can actually deal with both types of attributes, i.e., numerical and categorical. Interestingly, the whole NARM pipeline is usually less complex, because this does not need or demand any data preprocessing tasks as necessary by the original ARM, e.g., discretization. Consequently, most of the algorithms for dealing with NARM are more complex. The reason for this is that algorithms need to search for results within a much bigger search space, and, consequently, they must be tailored to consider this fact appropriately. Therefore, it is  natural  that most of these algorithms are based on stochastic population-based nature-inspired algorithms ~\cite{Fister2013BriefReview}, to which Evolutionary Algorithms (EAs) and Swarm Intelligence-based (SI) algorithms belong. 

Loosely speaking, we can confirm that EAs were the most appropriate for solving this task~\cite{minaei2013mining,tahyudin2017combination,Fister2018ARM}. However, there is also a noticeable trend of SI-based algorithms to be applied for solving the problem. Although the use of SI-based algorithms for NARM is still in the minority, some important works can also be found in this algorithm's family. The aim of this short review paper is to analyze the latest advances in solving the NARM using SI-based algorithms from the perspective of Five W`s and How: 
\begin{itemize}
    \item \textbf{What} are the features, differences and latest advances of SI-based algorithms for NARM?
    \item \textbf{Why} are the SI-based algorithms appropriate for NARM?
    \item \textbf{How} to use the SI-based algorithms for NARM?
    \item \textbf{When} to use the SI-based algorithms for NARM instead of EAs?
    \item \textbf{Who} use the SI-based algorithms for NARM?
    \item \textbf{Where} to use the SI-based algorithms for NARM?
\end{itemize}

A detailed literature review was conducted in order to obtain answers to these questions. The structure of this paper is designed as follows: Section~\ref{sec2} outlines briefly the fundamentals of the SI-based family of algorithms. Section~\ref{sec4} focuses on analysis of the algorithms designed for solving the NARM, while Section~\ref{sec3} gives a detailed review of SI-based algorithms for solving this problem. Further, this Section provides answers on the research questions set in this study. The paper is wrapped up with a short overview of performed work, followed by outlining the future challenges in Section~\ref{sec5}.
 
\section{Swarm intelligence in a nutshell}
\label{sec2}
SI-based algorithms~\cite{blum2008swarm} belong to the class of stochastic population-based nature-inspired algorithms. Besides EAs, they constitute a big group of so-called nature-inspired algorithms~\cite{Fister2013BriefReview}. The algorithms covered under this umbrella are inspired by the behaviors of different animal species, chemical processes, or even some other natural processes. A simple definition from paper~\cite{Fister2013BriefReview} elucidates their working principles: "\textit{SI concerns the collective, emerging behavior of multiple, interacting agents who follow some simple rules.  While each agent may be considered as unintelligent, the whole system of the multiple agents may show some self-organization behavior and thus can behave like some sort of collective intelligence.}"

In computer jargon we can simplify the definition by the following assertions: SI-based algorithms consist of individuals that travel within the search space during the simulated evolution. They are modified by the different variation operators, which can be rule-based or equation-based. These variation operators in SI-based algorithms are different for each of the algorithms. Only the fittest individuals according to the evaluation function survive in the evolutionary cycle.

In summary, Algorithm~\ref{simple-nia} presents a very trimmed pseudocode of the general nature-inspired algorithm.

\begin{algorithm}[htb]
\caption{Simple nature-inspired metaheuristic algorithm.}
\label{simple-nia}
\small
\begin{algorithmic}[1]
\STATE INITIALIZE\_population\_with\_random\_solutions;
\STATE EVALUATE\_each\_solution;
\WHILE {TERMINATION\_CONDITION\_not\_met} 
\STATE MODIFY\_solutions\_using\_specific\_variation\_operators;
\STATE EVALUATE\_modified\_solutions;
\STATE SELECT\_solutions\_for\_the\_next\_generation;
\ENDWHILE
\end{algorithmic}
\normalsize
\end{algorithm}

When we focus on line~4 of the pseudocode, where modification of the trial solution is performed, we can conclude that variation operators in EAs differ conceptually from those used in SI-based algorithms. While the former found their origins in Darwinian evolution (like crossover and mutation), the latter mimic some natural process, and move the trial solution in accordance with this.

\section{Overview of SI-based algorithm for NARM}
\label{sec4}
For the detailed literature review, we performed an extensive search through the major academic search engines, as well as databases, i.e Google Scholar~\footnote{https://scholar.google.com/}, Microsoft Academic Graph\cite{sinha2015overview,microsoft_academic_2019_2628216}, Scopus\footnote{www.scopus.com}, and IEEE Xplore\footnote{https://ieeexplore.ieee.org/Xplore/home.jsp}. Our search string consisted of a combination of the following keywords: ''\textit{numerical association rule mining}'', ''\textit{numeric association rule mining}'', ''\textit{swarm intelligence}'', and ''\textit{particle swarm optimization}''. When a hit appeared in Google Scholar, we also investigated all citations joined to this paper. Because of the many intersections of abstract on the one hand and content on the other hand, some studies cannot be included here.

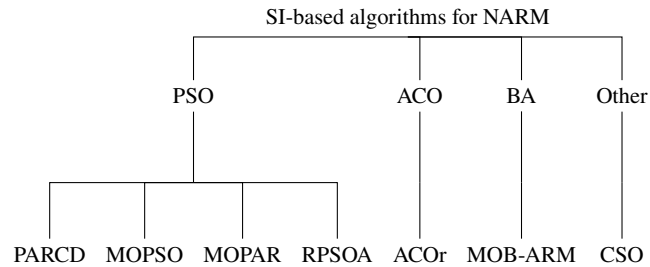
\begin{figure}
    \centering
\tikzset{edge from parent/.style=
{draw, edge from parent path={(\tikzparentnode.south)
-- +(0,0pt)
-| (\tikzchildnode)}},
blank/.style={draw=none}}
\begin{tikzpicture}
\matrix
{
\node{\Tree 
 [.{SI-based algorithms for NARM} 
    [.PSO [PARCD MOPSO MOPAR RPSOA ] ]
    [.ACO [{ACOr} ] ] 
    [.BA [ MOB-ARM ]]
    [.Other [ CSO ] ] ]};\\
};           
\end{tikzpicture}
\caption{A simple taxonomy of SI-based algorithms for NARM.}
\label{dendrogram}
\end{figure}

An archive was created based on the results of searching for papers including the aforementioned keywords. This served as grounds for analysis in the paper. At first, a simple taxonomy was deduced from the subjects of papers, and then these papers were analyzed deeply. Indeed, the simple taxonomy is depicted in Fig.~\ref{dendrogram},
which presents a cladogram, where the family of SI-based algorithms represents a root node. Thus, intermediate nodes denote the algorithms actually developed for solving the NARM, while the terminal nodes represent the variant of algorithms that are  presented in intermediate nodes. 

As can be seen from the Figure, the most popular SI-based algorithm for NARM is Particle Swarm Optimization (PSO)~\cite{kennedy1995particle}. Besides the PSO, there are also the other in this algorithm's family for solving the same problem, i.e.. Ant Colony Optimization (ACO)~\cite{dorigo2006ant}, Bat Algorithm (BA)~\cite{yang2011bat}, and Cat Swarm Optimization (CSO)~\cite{chu2006cat}. Each of these supports several variants of the original algorithms. 

The purpose of this paper is not to give a detailed overview of the nature-inspired algorithms. For more detailed information about nature-inspired optimization in general, readers are invited to read the following publication~\cite{cincin}. In the remainder of the paper, we are focused on analyzing the characteristics of the particular SI-based variants. 

\subsection{Particle Swarm Optimization NARM variants}
Probably, the first efforts using the PSO were performed by Alatas and Akin more than a decade ago, that proposed a Rough Particle Swarm Optimization for ARM (RPSOA)~\cite{alatas2008rough}. The RPSOA algorithm bases on the theory of rough patterns that use rough values defined with upper and lower bounds representing an interval of feasible values. 

One year later, the same authors proposed the multi-objective chaotic particle swarm optimization algorithm for classification rule mining~\cite{alatas2009multi}. Initially, the algorithm was not applied exclusively for numerical attributes, but also allowed combination of these attributes with categorical. An algorithm, called PSO for Numerical Association Rule Mining Problems with Cauchy Distribution (PARCD), ~\cite{tahyudin2017combination}, is another PSO variant that incorporates the Cauchy distribution employed by a mutation operator. This operator helps in avoiding the PSO algorithm trapping in local optima. The PARCD was also applied practically for predicting human positions~\cite{tahyudin2019predicting}. In the years since the introduction of the PARCD algorithm, an improved version was released by the same authors~\cite{tahyudin2019improved}.  
Multi-objective Particle Swarm Optimization for Association Rule Mining (MOPAR)~\cite{beiranvand2014multi} is a multi-objective PSO variant for mining the association rules in a single step. For dealing with numerical attributes rough values are used, containing the lower and upper bounds of intervals. A very interesting approach was presented by Kuo et al.~\cite{kuo2019multi}, where the MOPSO algorithm was used for solving the NARM problem. In order to deal with data in numerical form, the proposed algorithm includes the discretization procedure. This algorithm finds the best interval for each of the datasets without any effort by the user. Authors considered three criteria for association rule evaluation: confidence, interestingness and comprehensibility.

\subsection{Ant Colony Optimization NARM variants}
Besides the PSO, ACO is one of the first members of the SI-based algorithm’s family. In its original form it is dedicated for solving discrete optimization problems. Later, the Ant Colony Optimization for Continuous domains  (ACO$_{\text{R}}$) was developed, that also allows solving the continuous optimization problems by using the Gaussian function as a probability density function and its sampling. In line with this, authors Moslehi et al.~\cite{moslehi2011multi} proposed a multi-objective ACO$_{\text{R}}$ adjusted for dealing with NARM problems. Interestingly, the algorithm evaluates the solution quality as a weighted sum of even four criteria: support, confidence, interestingness, and amplitude. 

\subsection{Bat Algorithm NARM variants}
Actually, the BA has  rarely been applied to solving ARM problems. Some efforts exist~\cite{fister2019computational,Fister2017BatMiner} for solving the ARM with categorical attributes. However, the literature search revealed only one study that is directly related to NARM. Indeed, Heraguemi et al.~\cite{heraguemi2018multi} proposed a novel Multi-Objective Bat for ARM (MOB-ARM) applied to the NARM problems. The authors composed a fitness function of several quality criteria, as: support, confidence, comprehensibility, and interestingness.

\subsection{Other NARM variants}
The literature review also revealed the use of the CSO algorithm developed by Akyol et al.~\cite{akyol2016kedi}. Unfortunately, their paper is written in the Turkish language. Therefore, we just mention this algorithm in order to complete the overview without giving a detailed description of the algorithm's features.

\section{Analysis of algorithms for numerical association rule mining}
\label{sec3}
One of the biggest disadvantages of most algorithms for mining association rules is their handling with numerical attributes. Algorithms such as, for example, Apriori, require attributes in a dataset to be discretized before their use. Many times discretization can have a negative effect, due to the loss of information and noise entering into data. A new kind of algorithm for NARM appeared in order to overcome this drawback. Interestingly, stochastic population-based nature-inspired algorithms quickly turned out to be the most appropriate for this solving.

When solving the ARM using the stochastic nature-inspired population-based algorithms, two crucial components need to be adjusted:
\begin{itemize}
    \item the representation of solutions in the search space,
    \item fitness function evaluation.
\end{itemize}

The first component determines how the solution is encoded in the search space (genotype space). Typically, the majority of the observed stochastic nature-inspired population-based algorithms supports real-valued vector representation. Normally, the representation of solution in a search space does not reflect the same in the problem context (also phenotype space). Therefore, so-called genotype/phenotype mapping is employed for decoding the representation of a solution in a genotype space to its counterpart in the phenotype space.

The second component is devoted for estimating the quality of solutions in the phenotype space. This quality is calculated using the fitness function evaluation. In the following subsections, we describe the modifications of components in detail.

\subsection{Representation of solutions}
In general, algorithms for NARM have been using two approaches for representing individuals, i.e., Michigan and Pittsburgh. Using the former approach, each individual encodes a single association rule, whereas in the latter, each individual encodes a set of association rules. In the second approach, the ARM is divided into two steps: The first is necessary for finding the frequent item sets, while the second for generating the mined association rules based on their support and confidence. As showed by Srikant and Agrawal~\cite{srikant1996mining}, selecting intervals for numeric attributes is quite sensitive to the support and confidence values.

Several kinds of representation of individuals are distinguished in the Michigan approach. Here, we illustrated only those found in our review of papers. The first representation is used commonly in the NARM community~\cite{alatas2008rough,beiranvand2014multi,kuo2019multi}, and it encodes the association rule as a vector of attributes consisting of the following triples:
\begin{equation}
    (\langle \textit{ACN}_1, \textit{Lb}_1,\textit{Ub}_1\rangle,\ldots,\langle \textit{ACN}_n, \textit{Lb}_n,\textit{Ub}_n\rangle),
    \label{eq:rep1}
\end{equation}
where each triple contains three elements that determine: (1) presence of the attribute in antecedent ($A$), or consequent ($C$), or its absence ($N$) from the rule, (2) the lower bound and (3) the upper bound. In Eq.~\ref{eq:rep1}, $n$ denotes the maximum number of attributes. Mostly, the element $\mathit{ACN}_j$ is decoded according to the following relation:
\begin{equation}
    \mathit{o}_j=\begin{cases}
        \mathit{ACN}_j\leq 1/3, & \text{antecedent},\\
        1/3<\mathit{ACN}_j\leq 2/3, & \text{consequent},\\
        \mathit{ACN}_j> 2/3, & \text{not present}.
    \end{cases}
\end{equation}

The representation introduced by Alatas et al.~\cite{alatas2008rough} reduced the number of elements from the last one. As a result, the individual is represented as:
\begin{equation}
    (\langle\textit{AE}_1,\textit{AV}_1\rangle,\ldots,\langle\textit{AE}_n,\textit{AV}_n\rangle),
\end{equation}
where the AE$_j$ denotes attribute-existence part, and the AV$_j$ attribute-value part of the encoded attribute. Thus, the attribute-existence part is decoded as:
\begin{equation}
\mathit{o}_j=\begin{cases}  
        \mathit{AE}_j\leq.5, & \text{antecedent},\\
        \text{otherwise}, & \text{not present}.\\
    \end{cases}
\end{equation}
The attribute-value part determines the interval of feasible values according to complex calculation.

The authors of ACO for NARM, i.e., Moslehi et al.~\cite{moslehi2011multi}, proposed the following representation of individuals:
\begin{equation}
    (\langle \textit{ACN}_1, s_1,\sigma_1\rangle,\ldots,\langle \textit{ACN}_n, s_n,\sigma_n\rangle),    
\end{equation}
where each interval of feasible values is represented as $s_j\pm \sigma_j$. In this case, the meaning of the element $\mathit{ACN}_j$ is the same as in already mentioned representations.

An interesting representation was introduced by Heraguemi et al.~\cite{heraguemi2018multi}, where the real-valued individuals $\mathbf{x}_i=(x_{i,1},\ldots,x_{i,n+1})$ in BA algorithm are mapped to the phenotype space as:
\begin{equation}
    (\mathit{cp}_i,o_{i,1},\ldots,o_{i,n}),
\end{equation}
where $\mathit{cp}_i$ is the so-called cutting point determining the number of antecedent attributes belonging to the particular association rule, $o_{i,\Pi_j}$ denotes the index of the attribute arisen in the rule, and $\Pi_j$ determines the permutation of attributes. If this index is zero, the corresponding attribute is omitted from the association rule.

\subsection{Definition of the fitness function}
Fitness function plays an important role in evaluating the produced solutions. Therefore, researchers proposed different variations of these functions. Typically, the fitness function bases on various measures of significance and interest of the observed association rules. Usually, the following measures in the role of evaluation criterion can be found in the reviewed literature: support, confidence, comprehensibility, interestingness, and amplitude. 

The support of an association rule $X\Rightarrow Y$ determines how frequently the itemset $I=\{o_1,\ldots,o_n\}$ appears in transaction database $D=\{t_1,\ldots,t_M\}$. Formally, the measure is expressed as:
\begin{equation}
    \mathit{supp}(X\Rightarrow Y)=\frac{|X\cup Y|}{|D|}.
\end{equation}
where $|X\cup Y|$ designates the number of transactions containing antecedent $X$ and consequent $Y$. 

The confidence indicates the proportion of the transactions containing $X$ also contains $Y$. Mathematically, the measure is defined as:
\begin{equation}
    \mathit{conf}(X\Rightarrow Y)=\frac{\mathit{supp}(X\cup Y)}{\mathit{supp}(X)}.
\end{equation}
This equation can be interpreted also as an estimate of the conditional probability $P(E_Y|E_X)$ and expresses an accuracy that a definite association rule occurs in the transaction database.

Both mentioned measures are more or less statistical and say nothing about the quality of the association rule in the sense of reliability and coverage. Consequently, Gosh and Nath in~\cite{ghosh2004multi} defined comprehensibility and interestingness quality measures. According to these authors, the rules with less numbers of attributes in the antecedent are more comprehensible. In line with this, this measure is expressed as:
\begin{equation}
    \mathit{comp}(X\Rightarrow Y)=\frac{\log(1+|Y|)}{\log(1+|X\cup Y|)},
\end{equation}
where $|Y|$ is the number of attributes in the antecedent and $|X\cup Y|$ the number of transactions, for which it holds that, if $X$ then $Y$.

The interestingness measure is focused on discovering hidden information by extracting some interesting rules. Mathematically, it is defined as:
\begin{equation}
    \mathit{interest}(X\Rightarrow) Y)=\frac{\mathit{supp}(X\Rightarrow Y)}{\mathit{supp}(Y)}\cdot\frac{\mathit{supp}(X\Rightarrow Y)}{\mathit{supp}(X)}\cdot\left( 1-\frac{\mathit{supp}(X\Rightarrow Y)}{|D|}\right),
\end{equation}
that consists of three terms: The first denotes the probability based on the antecedent part, that second based on the consequent part, and the third based on the total number of transactions in the transaction database.

The amplitude is devoted for measuring the quality of the numerical association rules, and prefers the attributes with smaller intervals, as shown by Alatas et al. in~\cite{alatas2008modenar}. Formally, this measure is defined as:
\begin{equation}
    \mathit{ampl}(X\Rightarrow Y)=1-\frac{1}{n}\sum^{n}_{k=1}\frac{\mathit{Ub}_k-\mathit{Lb}_k}{\max(o_k)-\min(o_k)},
\end{equation}
where $n$ is the number of attributes in itemset $X\cup Y$, $\mathit{Ub}_k$ and $\mathit{Lb}_k$ are the upper and the lower bounds, while $\max(o_k)$ and $\min(o_k)$ limit boundaries of feasible values for attribute $o_k$.

As can be seen, the observed problem can be evaluated according to several objectives. It is actually a multi-objective problem in its nature. However, this kind of problems can be solved in several ways. In our review, we indicated especially two methods: (1) Weighted sum, and (2) Pareto dominance. The former transforms a Multi-Objective (MO) problem into single-objective. In other words, the evaluation function is expressed as:
\begin{equation}
    F=\sum^m_{k=1}w_k\cdot f_k,
\end{equation}
where $f_k$ is the $k$-th objective function, $w_k$ a weight assigned to this, and $m$ represents the number of objectives.

On the other hand, the Pareto dominance methods work with particular objectives simultaneously. These prefer so-called dominate solutions, where a solution $\mathbf{x}_1$ dominates a solution $\mathbf{x}_2$ if, and only if, $\mathbf{x}_1$ is strictly better than $\mathbf{x}_2$ for at least one of the objectives being optimized, and $\mathbf{x}_1$ is not worse than the $\mathbf{x}_2$ for all the other objectives~\cite{deb2001multi}. Interestingly, these solutions form the so-called Pareto Optimal Front (POF), from which users can select the best solutions according to their preferences.

\subsection{Discussion}
 A summary of the work in the domain of NARM is made in this Section. As can be seen from the performed study, modifications of especially two components are necessary for preparing the various stochastic nature-inspired population-based algorithms to solve the NARM, i.e., representation of individuals, and evaluation of fitness function. Typically, researchers use the Michigan and Pittsburgh approach for implementation of the former component, and either the weighted sum or multiobjective method for the second.

Summarizing this overview is illustrated in Table~\ref{fit-measures}, where each algorithm's variant representation approach (column ''Represent.'') is presented, together with the objectives managed (column ''Objective'') either, single- or multi-objective (column ''MO'') and corresponding reference (column ''Ref.''). 
\begin{table}
    \centering
    \caption{Quality measures that were used by each of the algorithms.}
    \label{fit-measures}
    \begin{tabular}{|l|l|l|c|c|c|c|c|c|r|}
    \hline
    \multirow{2}{*}{Algorit.} & \multirow{2}{*}{Variant} & \multirow{2}{*}{Represent.} & \multirow{2}{*}{MO} & \multicolumn{5}{c|}{Objective} & \multirow{2}{*}{Ref.}\\ \cline{5-9}
     & & & & Support & Confiden. & Interest. & Compreh. & Amplitud. & \\ \hline \hline
    \multirow{7}{*}{PSO} &\multirow{4}{*}{PARCD} & Michigan & Yes & & \greenbullet & & \greenbullet & & \cite{alatas2009multi} \\ \cline{3-10}
     & & Pittsburgh & Yes & & \greenbullet & \greenbullet & \greenbullet & & \cite{tahyudin2017combination} \\ \cline{3-10}
     & & Pittsburgh & No & & \greenbullet &  &  &  & \cite{tahyudin2019improved} \\ \cline{3-10}
     & & Pittsburgh & Yes & \greenbullet & \greenbullet & \greenbullet & \greenbullet & \greenbullet & \cite{tahyudin2019predicting} \\ \cline{2-10}

    & MOPAR  & Michigan & Yes & & \greenbullet & \greenbullet & \greenbullet & & \cite{beiranvand2014multi} \\ \cline{2-10}
    & RPSOA & Michigan & No\footnotemark[1] & \greenbullet & \greenbullet &  &  & \greenbullet & \cite{alatas2008rough} \\ \hline \hline
    ACO & ACO$_{\text{r}}$ & Michigan\footnotemark[2] & No & \greenbullet & \greenbullet & \greenbullet &  & \greenbullet & \cite{moslehi2011multi} \\ \hline \hline
    BA & MOB-ARM & Michigan & Yes\footnotemark[3] & \greenbullet & \greenbullet & \greenbullet & \greenbullet & & \cite{heraguemi2018multi} \\ \hline
    \end{tabular}
\end{table}

As can be seen from Table~\ref{fit-measures}, the majority of the observed SI-based algorithms for solving NARM employ the Michigan representation of individuals. On the other hand, the problem is handled as multi-objective, especially, in the recent publications. Weighted sum fitness function was characteristic only in the beginning of the NARM development at the end of the last decade. Interestingly, some of algorithms, like MOBARM, join four objectives into two pairs, where each pair is expressed as the weighted sum of objectives. Thus, the multiobjective problem with four objectives is reduced into a multiobjective problem with two objectives. The following five ARM measures serve typically for building the fitness function: support, confidence, interestingness, comprehensibility, and amplitude.

\footnotetext[1]{Weighted sum of objectives is employed.}
\footnotetext[2]{Value of solution and standard deviation conforms a representation of individual.}
\footnotetext[3]{Two objective functions are employed as a weighted sum of two measures.}

In the Introduction section of this paper, we set six research questions, where the answers to which would be addressed in the following paragraphs:
\begin{itemize}
    \item \textbf{What}: Probably the most important feature is handling numerical attributes in a similar way as the categorical attributes. Avoiding the discretization is  a very huge benefit.
    \item \textbf{Why}: SI-based algorithms are appropriate for dealing with data mining problems. This is also revealed by the fact that this family of algorithms are also applied to  other domains (e.g., for classification, regression, expert systems, etc.).
    \item \textbf{How}: Basically, each canonical algorithm needs modification of only two components (i.e., representation of individuals and fitness function evaluation).
    \item \textbf{When}: These algorithms can be applied to imbalanced data, large-scale datasets, and datasets with many attributes. Thus, feature selection is performed implicitly by using them.
    \item \textbf{Who}: Mostly, these methods are published by researchers in papers theoretically, while their implementations in the real-world are missing.
    \item \textbf{Where}: The method could be applied for building rule-based classifiers and expert systems.
\end{itemize}

Before performing the literature overview, we were sure that numerous papers exist for solving NARM, which are fully grounded on SI-based algorithms. However, the literature search revealed the opposite direction. According to the search records, only a few methods appeared in peer reviewed literature, when some records are linked even to  non-English languages. Therefore, we can conclude that this research area is slightly underestimated by the research community. For instance, according to Fig.~\ref{evolution-arm} we can see that ARM is a very popular topic, but solving NARM is still in the minority. Thus, we are confident that there is a pool of open research opportunities.

\section{Conclusions and future challenges}
\label{sec5}
As the volume of literature of the SI-based algorithms for NARM has still been rising in recent years, we aimed to provide a short but timely overview of the latest developments in the field. Different aspects of SI-based algorithms and their role in solving NARM have been summarized and highlighted. Practical applications where these methods can be utilized were also covered in this review.
Despite many successful applications of the SI-based algorithms to various domains, there are still a few key issues demanding more research effort. We must highlight that the SI-based family is enormous, and thus, we do not know how different algorithms behave in solving NARM problems. There is also a lack of theoretical analysis that should work in favor of SI-based algorithms for NARM. Finally, providing  software frameworks and releasing a source code of algorithms may help in spreading these approaches also into the real-world.

\end{document}